# A Logic-based Relational Learning Approach to Relation Extraction: the OntoILPER System

**Rinaldo Lima[1] • Bernard Espinasse[2] • Fred Freitas[3]**

[1] Departamento de Computação, Universidade Federal Rural de Pernambuco, Recife, Brazil
[2] Aix-Marseille Université, LIS-UMR CNRS, Marseille, France
[3] Centro de Informática, Universidade Federal de Pernambuco, Recife, Brazil

rinaldo.jose@ufrpe.br, bernard.espinasse@lis-lab.fr, fred@cin.ufpe.br

**Abstract** – Relation Extraction (RE), the task of detecting and characterizing semantic relations between entities in text, has gained much importance in the last two decades, mainly in the biomedical domain. Many papers have been published on Relation Extraction using supervised machine learning techniques. Most of these techniques rely on statistical methods, such as feature-based and tree-kernels-based methods. Such statistical learning techniques are usually based on a propositional hypothesis space for representing examples, i.e., they employ an attribute-value representation of features. This kind of representation has some drawbacks, particularly in the extraction of complex relations which demand more contextual information about the involving instances, i.e., it is not able to effectively capture structural information from parse trees without loss of information. In this work, we present OntoILPER, a logic-based relational learning approach to Relation Extraction that uses Inductive Logic Programming for generating extraction models in the form of symbolic extraction rules. OntoILPER takes profit of a rich relational representation of examples, which can alleviate the aforementioned drawbacks. The proposed relational approach seems to be more suitable for Relation Extraction than statistical ones for several reasons that we argue. Moreover, OntoILPER uses a domain ontology that guides the background knowledge generation process and is used for storing the extracted relation instances. The induced extraction rules were evaluated on three protein-protein interaction datasets from the biomedical domain. The performance of OntoILPER extraction models was compared with other state-of-the-art RE systems. The encouraging results seem to demonstrate the effectiveness of the proposed solution.

**Keywords** — Relation Extraction, Rule Induction, Information Extraction, Inductive Logic Programming, Relational Learning.

## 1. Introduction

Information Extraction (IE) is an important task in Text Mining and has the goal of both discovering and structuring information found in semi-structured or unstructured documents, leaving out irrelevant information [Jiang, 2012]. There are two major subtasks IE: *Named Entity Recognition* (NER), and *Relation Extraction* (RE). NER aims at identifying named entities from texts and classifying them into a set of predefined entity types such as person, organization, location, among others. Such entity types are the most useful for many application domains [Turmo et al., 2006].

   RE consists of two related subtasks: *detecting* and *characterizing* semantic relations among (named) entities in text. The first subtask is in charge of determining whether a relation between two given entities holds, whereas the second refers to the classification problem of assigning a relation type label to a particular relation instance. Most work on RE focuses on *binary relations*, i.e. relations between two entities, or arguments.

   NER and RE have been widely applied to extract useful information from the rapidly growing amount of publications in the biomedical literature. For instance, NER has been employed to recognize protein names (mentions), and it is regarded as of paramount importance in bioinformatics research [Li et al., 2015]. Another commonly addressed subtask in biomedical domain concerns the *Protein-Protein Interaction* (PPI) [Quian & Zhou, 2012] that, based on the results of a previous NER stage, aims at finding pairs of proteins within sentences such that one protein is described as regulating or binding the other.

---

[1] Corresponding author: Tel: +55 81 98405-2527; Rinaldo Lima. Email: *rinaldo.jose@ufrpe.br*;
ORCID Identifier : 0000-0002-1388-4824

Most of the state-of-the-art approaches to RE in general, and in PPI in particular, are based on supervised *statistical machine learning* methods, such as *feature-based* and *tree-kernels-based methods*. Such methods are based on a propositional hypothesis space for representing examples, i.e., they employ an *attribute-value* (propositional) representation that has some limitations, particularly in the extraction of complex relations, which usually demand more contextual information about the involving instances. In other words, this representation is not able to effectively capture structural information from parse trees without loss of information [Choi et al., 2013].

An alternative to such statistical machine learning approaches to RE is *the relational learning* approach which is able to generate classification models from rather complex data structures such as graphs or multiple tables [Furnkranz et al., 2012]. This work explores *Inductive Logic Programming* [Muggleton, 1991] as one of the major relational learning techniques.

Our working hypothesis is that given an automatic acquisition of substantial linguistic knowledge represented by first-order predicates, in combination with a logic-based relational learning technique able to induce expressive extraction rules; it is possible to generate highly accurate relation extraction models. Moreover, we argue two main ideas: the RE task should be performed by reasoning on the structural features of the examples, and a rich relational representation model of the sentences should define the structural features.

This paper presents OntoILPER, an RE system based on a supervised learner that induces symbolic rules for extracting binary relations between entities from textual corpora. OntoILPER takes profit of a rich relational representation of examples, overcoming the drawback of some current RE systems based on a less expressive hypothesis space for representing examples. Indeed, OntoILPER hypothesis space not only integrates information about node properties and relations in the form of relational features expressing structural aspects of examples, but can also be systematically explored by the learning component.

Another important contribution of OntoILPER to RE concerns the use of a domain ontology for both defining the relations to be extracted, and ontology population purposes. In the former, the domain ontology is employed as formal background knowledge, providing a highly expressive relational hypothesis space for representing examples whose structure is relevant to the RE task. In the latter, the relation instances extracted from text can be converted to the corresponding ontological instances in the domain ontology. This last task is also known as *Ontology Population* [Petasis et al., 2011].

Contrary to many RE systems, OntoILPER allows the integration of prior knowledge about the domain to be integrated into the induction of the extraction rules. Indeed, during both the searching and the rule induction processes, domain knowledge can be efficiently used as constraints to reduce search space. Empirical results on three RE datasets from the biomedical domain (PPI) suggest that OntoILPER is a valuable alternative RE method compared with some statistical learning methods for RE due to several reasons that we discuss here

The rest of this paper is structured as follows: Section 2 reviews both state-of-the-art RE systems based on supervised machine learning and fundamental concepts about ontologies and inductive logic programming addressed in this paper. In Section 3, we provide an overview of OntoILPER functional architecture, focusing on its major components. Section 4 reports and discusses the results of comparative experiments conducted on three datasets from the biomedical domain. Finally, Section 5 concludes this paper and outlines future work.

## 2. Supervised Machine Learning Approaches to Relation Extraction

This section focuses on related work concerning two supervised machine learning approaches to RE (i) the *statistical learning approach* and (ii) the *relational learning approach*.

### 2.1. Statistical Learning Approach

The supervised machine learning approach involves the generation of statistical classification models and is one of the most used in the RE community. Its central idea is to cast RE as a *classification problem,* i.e., to construct a classification model performing two distinct steps: *learning* and *prediction*. During learning, the algorithm builds a model from labeled data that is able to separate the training data into two classes, say *A* and *B*. This is also called *binary classification*. During prediction, the learned model is applied to unlabeled instances in order to classify them into one of the *A* and *B* classes. Two major types of supervised machine learning approaches to RE are *Feature-based* and *Kernel-based* approaches.

*2.1.1 Feature-based approaches to RE*

Feature-based approaches to RE build classification models by first transforming relation examples into numerical vectors representing several kinds of *features* and then, a machine learning technique is employed, such as Support Vector Machine (SVM) [Joachims, 1999], to detect and classify the relations examples into a predefined set of relationship types. Such approaches achieve state-of-the-art performance results by employing a large number of linguistic features derived from lexical knowledge, entity-related information, syntactic parsing trees, and semantic information [Kambhatla, 2004] [Zhou et al., 2005] [Giuliano et al., 2006] [Li et al., 2015] [Muzaffar et al., 2015]. The utilization of thousands of features is computationally burdensome and does not scale well with increasing amount of data. Furthermore, it is difficult for the feature-based approaches to effectively capture structured parse tree information, which is critical for further performance improvement in RE [Zhou et al., 2005].

*2.1.2 Kernel-based approaches to RE*

Kernel-based approaches to RE are based on kernel functions, or simply kernels, that define the inner product of two observed instances represented in some underlying vector space. Kernel functions are often regarded as a measure of similarity between two input vectors that represent examples in a transformed space using the original attribute set. The following two major types of kernels have been explored in RE [Jiang, 2012]:

*Tree-based kernels* are based on common substructures containing two entities in order to implicitly exploit structured features by directly computing the similarity between two trees, as it is done in [Culotta and Sorensen, 2004] [Airola et al., 2008] [Quian & Zhou, 2012] [Ma et al., 2015]. Tree-based kernels explore various structured feature spaces by processing parsing trees in order to capture syntactically structured information from the examples. Tree kernels can achieve comparable or even better performance than feature-based ones, largely due to their distinctive merit in capturing, to some extent, the structural information of relation instances. However, there exist two main problems in applying tree-based kernels in RE. The first one is that the subtrees in a tree kernel computation are context-free; therefore, they do not consider the context information outside the target subtree containing two argument entities [Zhou et al., 2007]. The second problem concerns the choice of a proper tree span in RE, i.e., the tree span relating the subtree enclosed by the shortest path linking two involving entities in a parse tree [Zhang et al., 2006].

*Composite kernels* result from the combination of different kernels [Miwa et al., 2010] [Tikk et al., 2010]. Composite kernels are mainly used when it is difficult to include all kinds of features into a single kernel, i.e., they can integrate the advantages of feature-based and tree kernel-based ones. Zhao and Grishman (2005) defined several feature-based composite kernels in order to integrate diverse features. In [Zhang et al. 2006], the authors propose a composite kernel that combines the convolution parse tree kernel with the entity feature kernel. More recently, Choi et al. (2009) introduced a composite kernel that integrates different kinds of lexical and contextual features by expanding an existing composite kernel. They extended the syntactic features with a range of lexical features for achieving more accurate extraction results. Previous investigation [Choi et al., 2009] [Jiang, 2012] revealed that composite kernels achieve better performance than a single syntactical tree kernel. This means that entity type information can be combined with structural (syntactic) features into a single kernel function. The disadvantage of composite kernels resides in the fact that the comparison is performed only on the basis of the sentence component information of each node [Jiang, 2012].

**2.2. Relational Learning Approach**

The supervised learning approach referred to as Relational Learning generates classification models from complex data structures (graphs or multiple tables) [Fürnkranz et al., 2012]. In this approach, one of the most widely-used learning techniques is *Inductive Logic Programming* (ILP) that employs first-order predicates as a uniform representation language for examples, background knowledge (BK), and hypotheses [Lavrac and Dzeroski, 1994]. Additionally, available expert knowledge can be used as further BK during learning in ILP, which increases the expressiveness of the hypothesis space.

In the remainder of this section, we first introduce two major components of the ILP-based systems for RE discussed in this work: ILP and ontologies. Then, some current ILP-based RE systems are presented followed by a qualitative comparison among them.

*2.2.1. Inductive Logic Programming*

ILP is theoretically settled at the intersection of Inductive Learning and Logic Programming. From inductive machine learning, ILP inherits the development of techniques to induce hypotheses from observations. From Logic Programming, it inherits its representation formalism and semantics.

As a supervised learning technique, ILP that aims at learning an *intentional description (or hypothesis H)* of a certain target *predicate*, based on some BK, and two sets of positive ($E^+$) and negative examples ($E^-$), generally represented by clauses with no variable, i.e., *ground clauses*. The induced hypothesis H is represented as a finite set of conjunctions of clauses H of the form $H \leftarrow h_1 \wedge \cdots \wedge h_k$ where each $h_i$ is a non-redundant clause, that entails all the positive and none of the negative examples [Muggleton, 1995].

One of the reasons for the ILP success is the readability of the induced models. In addition, it has the capability of learning from structural or relational data, therefore taking profit of domain knowledge given as BK. Another interesting advantage is that it can represent, using first-order logic, more complex concepts than traditional attribute-value (zero-order) languages [Furnkranz et al., 2012].

*2.2.2 Ontologies*

In one of the most cited definitions of ontology, Gruber asserts that ontology is an explicit specification of a conceptualization [Gruber, 1993]. Ontologies are representations of formalized knowledge that can be processed by a computer in a large number of tasks, including communication and interoperability (using ontology as a common vocabulary), communication and reasoning of intelligent agents. In practical terms, ontologies encompass definitions of concepts, properties, relations, constraints, axioms and instances on a certain domain or a universe of discourse. In addition, they allow for the reuse of domain knowledge, making domain assumptions explicit. The interest of using ontologies in the IE process has been demonstrated by several researchers. Ontologies have been used for multiple purposes: capture knowledge on a given domain [Nedellec & Nazarenko, 2005], for processing information content [Karkalesis et al, 2011], and reasoning [Wimalasuriya & Dou, 2009], just to name a few.

*2.2.3 ILP-based systems for RE*

Kim et al. (2007) propose a RE system based on the Aleph[2] ILP system. It performs text preprocessing using NER, Part-of-Speech (POS), chunking analysis, and grammatical function assignment (subject, object, time, etc.) provided by the Memory-based Shallow Parser (MBSP)[3]. This system was evaluated using a set of sentences from the PRINTS database, a protein family database. The evaluation task concerns extracting relation instances between proteins and other biological entities, including disease, function, and structure. Their reported results achieved 75% precision, but with a low recall of less than 30% for two of the three evaluated datasets.

The RE system proposed by Horvath et al. (2009) considers dependency trees as relational structures composed of binary predicates representing the edges in a graph. In their solution, the text preprocessing component is based on both the GATE framework [Cunnighan et al., 2002] and the Stanford parser [De Marneffe and Manning, 2008]. They also employed WordNet [Fellbaum, 1998] as a semantic resource for obtaining hypernym relations from two given entities found in text. In addition, the authors assume a partial order on the set of unary predicates which are defined by a hierarchy between entities, e.g., the unary predicate *Person(X)* is more general than the *Physicist(X)* predicate. Applying the Least General Generalization (LGG) [Plotkin, 1971] technique, they generate a set of rules expressed as non-recursive Horn clauses satisfying some criteria of consistency, e.g., all rules must cover a minimum number of positive examples. Then, the generated rules are used for constructing a binary vector of attributes for each example. Finally, the resultant vectors are employed for training a SVM classifier.

Smole et al. (2012) propose an ILP-based system that learns rules for extracting relations from definitions of geographic entities in the Slovene language. Their system is used as a component in a spatial data recommendation service. The authors focus on the extraction of the five most frequent relations ("isA", "isLocated", "hasPurpose", "isResultOf", and "hasParts") present in 1,308 definitions of spatial entities. Their Natural Language Processing (NLP) component is based on the Amebis Slovene POS tagger. The authors implemented a tool for chunking detection in Slovene that takes as input the text already tagged by Amebis. The learning component in this system is based on the Progol ILP system [Muggleton, 1995].

Kordjamshidi et al. (2012) provide a relational representation to the so-called Spatial Role Labeling (SpRL) problem introduced in [Kordjamshidi et. al, 2011]. This problem concerns the extraction of generic spatial relations from texts. The main tasks treated in a SpRL problem are (i) the identification of the words describing spatial concepts; and (ii) the classification of the role that these words play in the spatial configuration. The authors employed kLog [Frasconi et al., 2012], a framework for kernel-based relational learning that uses graph kernels. kLog can also take profit of BK in the form of logic programs. The authors employed the Charniak Parser [Charniak and Johnson, 2005], for POS tagging and dependency parsing.

---

[2] The Aleph Manual. http://www.cs.ox.ac.uk/activities/machlearn/Aleph/aleph.html
[3] Memory-based Shallow Parser for Python. http://www.clips.ua.ac.be/ctrs/memory-based-shallow-parser-for-python

Nédellec et al. (2008*)* present the Alvis system, a RE system that extracts relations between biological entities. Alvis provides a semantic analysis based on the Ogmios NLP framework [Nazarenko et al., 2006] which performs several NLP subtasks, including NER of biological entities, POS tagging, syntactic parsing, and semantic mapping to biological domain ontologies. Alvis is based on LP-Propal, an ILP-based learning component proposed in [Alphonse and Rouveirol, 2000]. LP-Propal takes the annotated corpus as input for inducing extraction rules suitable to tag semantic relations found in the domain ontology. Alvis heavily depends on terminology dictionaries for identifying biomedical entity instances in texts. This seems not adequate for very active domains that produce documents introducing new named entities, due to the considerable effort required to keep such terminology dictionaries fully updated.

*2.2.4 Qualitative Comparison of ILP-based RE systems*

Tab. 1 summarizes the characteristics of the ILP-based RE systems presented above according to the following dimensions: (i) IE task performed, either NER or RE, (ii) NLP subtasks performed in text preprocessing, (iii) NLP tool(s) used, (iv) linguistic or semantic resources used, and (v) the ILP learning component, (vi) evaluation dataset(s), and (vii) use of ontologies.

According to Tab. 1, Alvis and OntoILPER are the only systems performing both NER and RE. For the other systems, it is assumed that the NER problem is already solved, i.e., they take as input a corpus containing pre-annotated named entities.

Concerning the natural language tools used, GATE provides a comprehensive set of NLP tools in just one software package in Horvath's system. The Ogmios framework is another comprehensive NLP platform that offers several natural language tools but it is tailored to the biomedical domain, while GATE is domain-independent. In addition, it is noticeable the trend of ILP-based IE systems that rely on deeper NLP tasks such as full dependency parsing [De Marneffe and Manning, 2008], and SRL [Harabagiu et al., 2005]. Again, this is not surprising because previous work on dependency parsing and SRL has proved to be very beneficial to IE [Jiang and Zhai, 2007] [Harabagiu et al., 2005].

Considering external resources, such ontologies and semantic thesauri used in the IE process, only Hovarth et al. (2010) take profit from the WordNet.

Each studied system employs a different implementation of an ILP-based learning component. In addition, from the experimental setup point of view, half of them carried out significant experiments using several corpora either of the same domain or from different domains (Hovarth, Nédellec, and OntoILPER). Finally, concerning the use of ontologies, Nédellec's and OntoILPER systems are the only ones that take profit of them.

Almost all of the surveyed ILP-based systems assume that NER is already solved, i.e., they take profit of the pre-annotated named entities from the input corpus. This assumption limits their application to other corpora in which none of the named entities are already indicated. On the contrary, OntoILPER can effectively perform both NER and RE tasks as demonstrated by empirical results reported in [Lima et al., 2014a, 2014b], whereas Section 4 discusses comparative assessment results of OntoILPER with other state-of-the-art RE systems on several biomedical datasets. Moreover, contrarily to the majority of the systems in Tab. 1, OntoILPER can be considered as an OBIE system as it offers all the benefits of the synergy between the ILP-based learner and the domain ontology: the former is able to generate symbolic extraction rules, while the latter can be fully exploited by the OBIE process for generalization purposes.

To sum up, it was shown that the kernel-based approach using attribute-value representation has been more popular than the relational approach to RE. One possible reason is that the kernel-based approach to RE has obtained state-of-the-art performance on several shared tasks on both domains of newswire and biomedical [Li et al., 2013 e Ma et al., 2015]. On the other hand, the richer relational representation of the hypothesis space used by ILP-based approaches seems to be the more natural way of dealing with graph representation of the parsing of natural language sentences due to its advantages as already discussed in the introductory section of this paper.

Another important aspect exposed by our investigation is that there is a lack of a deeper experimental evaluation of ILP-based solutions to RE using several benchmark datasets either of the same domain or different domains for achieving more significant findings and conclusions.

In order to alleviate this gap, we propose the ILP-based RE system OntoILPER which takes profit of BK in terms of ontological elements of the domain under study. Moreover, OntoILPER was evaluated on three challenging benchmark RE datasets from the biomedical domain. More precisely, in the experimental section of this work, we conducted comparative assessments using state-of-the-art kernel-based RE systems instead of ILP-based RE systems which are certainly more closely related to OntoILPER. The justification for such a decision is based on the fact that the selected kernel-based RE systems employed the same experimental setting and the same publicly available benchmark RE datasets, which allowed for a more direct and fair comparison with OntoILPER; contrarily to the almost all of the ILP-based RE systems that were evaluated using tailored or non-publicly available corpora.

**Table 1**. Summary of the ILP-based RE systems

| Reference | IE task | Preprocessing task(s) | NLP tool used | Semantic resources used | ILP System | Evaluation dataset | Use of ontology in IE |
|---|---|---|---|---|---|---|---|
| Kim et al. (2007) | RE | POS, NER, Chunking, Grammatical functions | MBSP | None | Aleph | Function (1268)[a] | No |
| Nédéllec et al. (2008) | NER, RE | POS, NER, Gazeteers, Dep. parsing, Semantic Mapping | Ogmios Framework | None | LP-Propal | LLL | Yes |
| Hovarth et al. (2010) | RE | POS, Dep. Parsing | GATE and Standford Parser | WordNet | LGG | ACE 2003 | No |
| Smole et al. (2012) | RE | POS, chunking[b], Manual SRL | Amebis | None | Progol | 1308 definition sentences | No |
| Kordjamshidi et al. (2012) | RE | POS, Dep. Parsing, SRL | Charniak Parser and LHT tool | None | KLog | 1213 sentences | No |
| OntoILPER | NER, RE | POS, NER, Dep. Parsing, Chunking | CoreNLP and OpenNLP | None | ProGolem | LLL, HPRD50, IEPA[c] | Yes |

[a] Two other datasets were used *Disease* and *Structure* with 777 and 1159 sentences, respectively.
[b] POS tagger and Chunker for the Slovene language
[c] Well-know datasets from the biomedical domain presented in Section 4.1

## 3. OntoILPER: An ILP-based System for RE

OntoILPER, the RE system proposed in this paper, integrates both a domain ontology and an ILP learning component that induces symbolic extraction rules. These rules are used to classify binary relations involving entity instances. For that purpose, OntoILPER employs a relational model of sentences based on syntactical dependencies among linguistic terms (words, phrases). Such dependencies are regarded as *logical predicates,* which can be exploited by an automatic induction process. OntoILPER is based on the principle that the establishment of a relationship between two entities in the same sentence can be obtained by a path between them in a dependency graph, encoding grammatical relations between words or phrases.

Contrarily to the majority of the ILP-based IE systems discussed in Section 2.2, OntoILPER is centered on the idea of making domain knowledge explicitly available via an ontology, bringing two desirable benefits to the IE process: ontologies can offer new opportunities for IE systems, ranging from using them for storing the extracted information to using reasoning for improving IE tasks; and the portability of the IE system may be enhanced by adapting the system behaviour via changes in the ontologies.

Another important aspect that differentiates OntoILPER from related work (cf. Section 2.2) resides on our more rigorous experimental methodology which includes comparative evaluations on several standard datasets for RE. In fact, considering the experimental methodology adopted by related work, none of the studies carried out substantial experiments using several corpora either of the same domain or from different domains. For instance, in Seneviratne et al. (2012), only one relation type was evaluated on a set of 13 web pages. By contrast, [Hovarth et al., 2010] and [Nédéllec et al., 2008] were the only works that used standard datasets in their experimental assessments, but no comparative assessment is provided.

In the remainder of this section, we first present an overview of the OntoILPER system by describing its global extraction process, architecture, and components. Then, we focus on its major components involved in the RE process. Finally, we discuss the role played by the domain ontology in our solution.

## 3.1 OntoILPER Overview

OntoILPER is composed of several components arranged in a modular pipelined architecture as shown in Fig. 1. The core component integrates an ILP-based system that generates extraction rules in symbolic form. Such rules may be easily understood by a knowledge engineer, which can refine them in a later stage of the rule induction task, aiming at improving the whole extraction process. Moreover, extraction symbolic rules can be automatically converted into other rule formalisms, such as SWRL [Horrocks et al., 2010].

### 3.1.1 Global RE Process

In OntoILPER, the RE process is performed in two distinct phases, as illustrated in Fig. 1. First, in the *Extraction Rule Induction* phase, a theory (a set of rules) is induced by a general ILP system from an annotated learning corpus given as input. Then, in the *Extraction Rules Application* phase, the set of the induced rules is applied on the candidate instances (entities and relations) present in an unseen document in order to classify them before populating the domain ontology.

In both phases mentioned above, a previous preprocessing stage takes place in which several NLP tools are employed for generating a linguistically rich corpus annotation. This is followed by an automatic graph-based representation of the examples as first-order logic predicates.

### 3.1.2 Overview of OntoILPER Architecture

Fig. 1 shows OntoILPER major architectural components as darker boxes. Each component performs a specific subtask and is briefly presented next.

**Natural Language Processing Component.** OntoILPER integrates both the Stanford CoreNLP[4] for carrying out the following sequence of NLP subtasks: sentence splitting, tokenization, Part-of-Speech (POS) tagging, lemmatization (which determines the base form of words), NER, and dependency parsing; and the OpenNLP[5] tool, for the chunking analysis. For both verbal and nominal chunks, we consider its rightmost token as the head element. All of the linguistic analysis listed above is serialized in an XML file that contains a rich annotated version of the input corpus.

**Graph-based Sentence Representation and Graph Reduction Component.** This component is responsible for two closely related processing tasks in OntoILPER pipeline: the graph-based representation of sentences, and the graph reduction.

**Background Knowledge Generation Component.** This component automatically generates and represents relevant features from an annotated set of documents and a domain ontology as input. The generated features are converted into BK implemented as a Prolog factual base.

**ILP Rule Learning Component.** It relies on ILP to induce Horn-like extraction rules from the annotated examples. The extraction rules follow the same syntax of a Prolog predicate.

**Rule Application Component.** It applies the induced rules on the Prolog factual base generated from new documents not used in the rule learning phase. As a result, new instances of relations are identified and extracted.

**Ontology Population Component.** This component stores the new instances of relations as instances of the domain ontology classes.

The components (2)-(5) are further described in the remainder of this section, as well as, the role that the domain ontology plays in OntoILPER.

## 3.2 Graph-based Sentence Representation and Graph Reduction Component

### 3.2.1. Graph-based Model for Sentence Representation

OntoILPER employs a *relational (graph-based) model* of sentences based on both structural and property features that altogether describe mentions of entities and relations. Such features are regarded as *relational* or *logical predicates*. In this relational model for sentence representation, a binary relationship can be specified between conceptual entities (instances of classes and relations): each major phrasal constituent (nominal and verbal phrases) in a sentence is regarded as a candidate instance for extraction. In other words, all phrases expressing individual tokens or phrases are potentially referencing real-world concepts defined

---

[4] Stanford Core NLP Tools. http://nlp.stanford.edu/software/corenlp.shtml
[5] OpenNLP. http://opennlp.apache.org

by the domain ontology. Thus, the relational model can represent any syntactic structure of a given sentence
*S* as the graph mapping *G* applied to a sentence into a tuple of binary relations: *G: S → tuples of relations*.

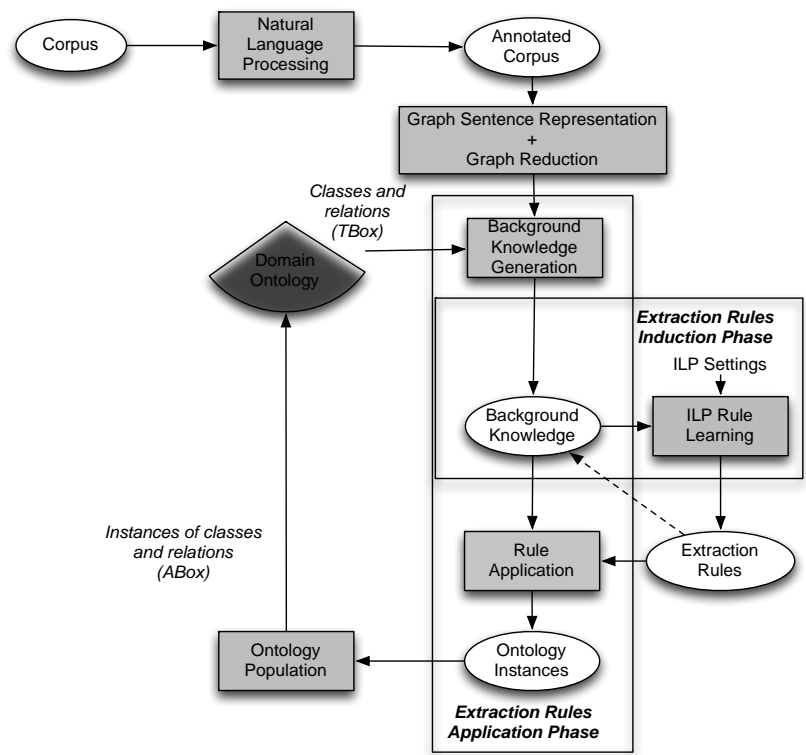

**Figure 1.** Overview of OntoILPER

This relational model for sentence representation is based on the principle that the establishment of a relationship between two entities in the same sentence can be obtained, for instance, by a path between them in this graph, which encodes both morpho-syntactical attributes of individual words, and semantic relations between phrases constituents [De Marneffe and Manning, 2008].

In what follows, we describe the two linguistic analyses that compose the proposed graph-based model for sentence representation.

**(i) Chunking Analysis.** Chunking analysis is used to define entity boundaries, and the *head (core)* constituents of nominal, verbal and prepositional phrases. For instance, consider the sentence "*Myron Kandel at CNNfn Newsdesk in New York*". Fig. 2 shows the head tokens of this sentence obtained after a performing chunking analysis on it. Usually, verbal phrases are possible candidates for relations, whereas nominal ones can represent an entity or an instance of a class.

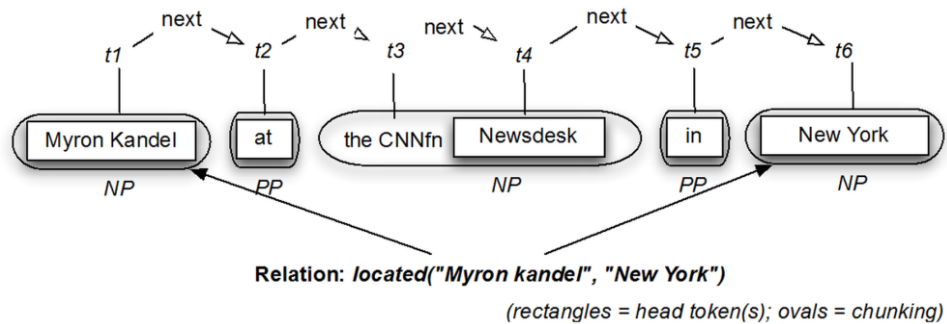

**Figure 2.** Chunking analysis and head tokens of the sentence

**(ii) Dependency Analysis.** The dependency analysis [Kruijff, 2002] consists of generating the typed dependencies parses of sentences from phrase structure parses and produces a *dependency graph* [De Marneffe and Manning, 2008]. A dependency graph is the result of an all-path parsing algorithm based on a *dependency grammar* [Jiang and Zhai, 2007] in which the syntactic structure is expressed in terms of dependency relations between pairs of words, a *head* and a *modifier*. The set of derived relation according to the dependency grammar defines a *dependency tree*, whose root is usually the main verb of the sentence.

We have adopted the typed dependencies proposed in [De Marneffe and Manning, 2008], also called *Stanford dependencies*. Typed dependencies and phrases structures are different ways of representing the inner structure of sentences, in which a phrase constituent parsing represents the nesting of multi-word constituents, whereas a dependency parsing represents dependencies between individual words. In addition, a typed dependency graph labels dependencies with grammatical relations, such as *subject* or *direct object*. Fig. 3 shows the dependency graph of a sentence with typed dependencies denoted by directed edges.

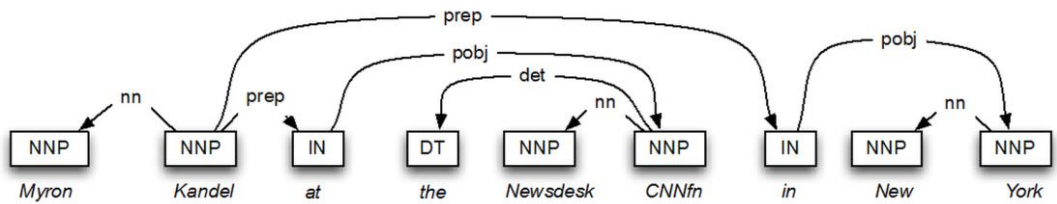

Figure 3. Dependency graph of the sentence "*Mary Kandel at Newsdesk CNNfn in New York*"

Fig. 4 shows an instance of the relational model of the sentence introduced above which is obtained by integrating: (i) a dependency analysis with *collapsed dependencies* (e.g. *prep_on*) generated by the Stanford dependency parser; (ii) a *chunking analysis* (head tokens in bold), i.e., the sequencing of tokens in a sentence (*NextToken* edges), (iii) *morpho-syntactic features* as node attributes (arrows in gray color), and (iv) named entity attributes.

According to the example shown in Fig. 4, one can identify a set of binary relations or predicates, including *det(Newsdesk, the), nn(Newsdesk, CNNfn), prep_at(Myron-Kandel, Newsdesk), prep_in(Myron-Kandel, New-York)*.

This collection of binary relations and their arguments can be further enriched with additional constraints on the types of the arguments. Such additional binary relations are used by OntoILPER induction process to link terms in a sentence with classes and relations from the domain ontology. For illustrating this point, consider a target relation to be learned, e.g., *located(X, Y)* as shown in Fig. 4, then the first argument $X$ should be an instance of the *Person* class, while the second argument $Y$ should be an instance of the *Location* class in the domain ontology.

Therefore, in this proposed relational model, instances of classes and relations can be viewed as nodes and edges, respectively. Moreover, each node can have many attributes, including ontology class labels.

From the perspective of the RE task in OntoILPER, the grammatical dependencies between words in a sentence are considered as relational features that can be exploited in the induction of symbolic extraction rules from sentences. Additionally, OntoILPER approach to RE rests on the hypothesis that, when learning about properties of objects represented by rich relational models, feature construction can be guided by the structure of individual objects and their relationships to other objects.

*3.2.2. Graph Reduction*

We proposed in [Lima et al., 2013] several heuristic-based rules for reducing graphs representing rich sentence annotation as shown in Fig. 4. More specifically, our graph reduction strategy replaces dependency graphs of sentences with smaller versions of them. This graph reduction step is rather *entity-oriented*, in the sense that it seeks to preserve the minimal relevant contextual information around entities in a dependency graph. The key idea is to speed up the learning phase by applying several rules for graph reduction that restricts the hypothesis space and increases recall by enabling more general shorter graphs.

The rules have the form $R_i : \{C_i\} \rightarrow \{A_i\}$, where $C_i$ denotes the conditional part which is mainly defined by constraints on node attributes including its POS tags, its type of outgoing/incoming edges, its parent nodes, etc.; and $A_i$ is a series of operations applied on the matched nodes to remove some edges from the graph. We refer the reader to [Lima et al., 2014a] for the definition of such rules.

As reported in our previous work [Lima et al., 2014a], this graph reduction strategy enables the discovery of more general extraction rules, as they proved to be very useful improving OntoILPER performance on two domains: biomedical and news. Another advantage resides in the fact that the simplified graphs reduce search in the hypothesis space, resulting in less learning time.

*3.2.3. Relational Representation vs. Vector Representation.*

In *propositional* machine learning, all the examples are represented by a single table, in which a row denotes an example, and a column denotes an attribute of the example. In this tabular or *vector-based* representation, the incorporation of expert knowledge about a given domain is usually done by adding new columns as function of other data columns.

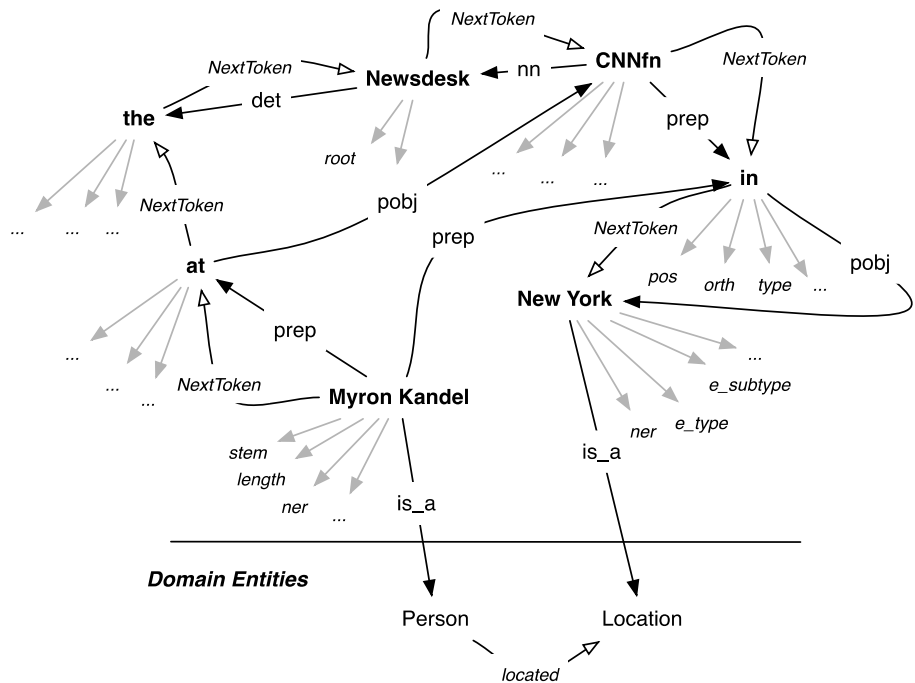

**Figure** 4. Example of the graph-based model of the sentence: "*Myron Kandel at the Newsdesk CNNfn in New York*"

The kernel-based methods for RE [Giuliano et al., 2007] [Jiang and Zhai, 2007] [Li et al., 2015] [Muzaffar et al., 2015] take as input structural representation of sentence parsing trees and convert them into features in a vector representation model. This conversion is usually performed by applying similarity functions on the sentence parsing trees. As a result, part of the relational knowledge, i.e., the structural information is lost in this transformation process [Fürnkranz et al, 2012] [Jiang, 2012].

Another limitation of the vector representation of examples resides in the restriction of having a unique representation format for all the examples, i.e., one feature is created for each element in the domain, and the same feature is used for characterizing all examples under consideration. In general, this results in a very sparse data table because most of the attributes will contain null values, due to the difference among the examples. Yet, Brown and Kros (2003) pointed out that this data sparseness problem is even more critical when deep knowledge is explored, which can cause serious problems for propositional machine learning algorithms.

By contrast, in OntoILPER, each example is represented independently of the others. Thus, the data sparseness problem for representing the examples is highly reduced [Fürnkranz et al, 2012]. Thereby, the above limitations are alleviated by employing first-order formalism for representing both BK and examples. This enables that many sources of information, either propositional or relational in nature, to be effectively represented without the drawbacks of the vector representation mentioned above. Moreover, the ability to take into account relational BK and the expressive power of the language of the induced rules are OntoILPER distinctive features.

### 3.3. Background Knowledge Generation Component

The primary goal of the *BK Generation Component* is to identify and extract relevant features from the relational model presented in the previous section. Guided by the domain ontology, this component converts the generated features into a *Prolog factual base* which is used as input by the ILP-based learner.

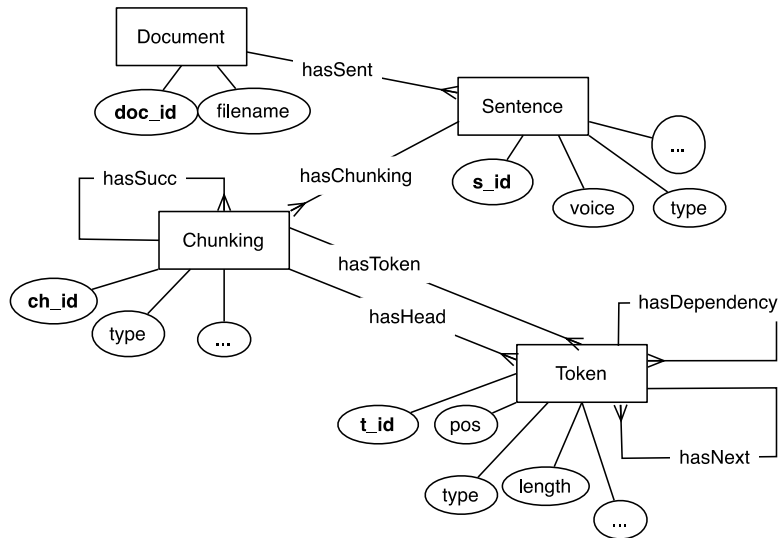

**Figure 5** Entity-Relationship model used for BK generation in OntoILPER.

In addition, the generated features can be structured by the *Entity-Relationship* (ER) diagram depicted in Fig. 5. This ER model shows how domain objects, including documents, sentences, phrases, and tokens are structured in the knowledge base. In combination with the domain ontology, the structure defined by the ER model can be used for guiding the generation of binary predicates and their arguments, both enriched with additional constraints on their types. Other binary predicates are also generated and they aim at linking terms in a sentence with classes and relations from the domain ontology. The underlying idea here is that when learning about objects in RE, feature construction should be guided by the structure of the examples.

In what follows, we described the proposed features in detail as well as user-defined BK.

**Structural and Attributive Features**. There exist four main groups of features in OntoILPER, namely:
- **Lexical features** which concern word, lemma, length, and general morphological type information at the token level.
- **Syntactic features** which consist of *word POS tags*; *head word* of nominal, prepositional or verbal chunk; *bi-grams* and *tri-grams* of consecutive POS tags of words as they appear in the sentence[6]; *chunking* features that segment sentences into noun, prepositional, and verb groups providing *chunk type* information (nominal, verbal or prepositional), *chunk head word*, and its *relative position* to the main verb of the sentence.
- **Semantic features** include the recognized named entities in the text preprocessing phase, and any of the additional *entity mention* feature provided by the input corpus. For instance, in the TREC dataset, each annotated entity has its *entity mention type* (person, organization, or location).
- **Structural features** consist of the structural elements connecting all the other features in the graph-based model for sentence representation. They denote (i) the *sequencing of tokens* which preserves the token order in the input sentence; (ii) the *part-whole* relation between tokens and the chunk containing them, i.e., the tokens are grouped in its corresponding chunk; (iii) the sequencing of chunks is represented by edges between their head tokens; and (iv) the *grammatical dependency* between two tokens in a sentence according the typed dependencies between words given by the Stanford dependency parser.

Since Prolog is used as the representation language of the examples in OntoILPER, domain entities, relations, and all the types of features mentioned above are converted to the corresponding Prolog predicates. We illustrate a complete set of the features presented above with the instance of the *Person* class, "*Myron*" in Tab. 2. For most of the predicates in Tab. 2, the first-order logic representation of the features is straightforward: a *unary predicate* in Prolog denotes identifiers, whereas *binary predicates* correspond to attribute-value pairs, and relations, e.g., *rel(arg$_1$, arg$_2$)*. Differently from other machine learning approaches that employ feature vectors for representing *context windows* (*n* tokens on the right/left of a given word *w* in a sentence), we employ the binary predicate *next/2* which relates one token to its immediate successor in a sentence, as shown in Tab. 2.

---

[6] We have also experimented with 4-grams, but bi-grams and tri-grams achieved better results in our preliminary experiments

**Table 2.** Prolog predicates describing the token "Myron" (t_1)

| Group | Prolog Predicates | Meaning |
|---|---|---|
| **Corpus entities** | doc(d_1) | d_1 is a document identifier |
| | sent(s_1) | s_1 is a sentence identifier |
| | chunk(ck_1) | ck_1 is a chunk identifier |
| | token (t_1) | t_1 is a token identifier |
| **Lexical features** | t_stem (t_1, "Myron") | token t_1 stemming is "Myron" |
| | t_length (t_1, 5) | token t_1 has length of 5 characters |
| | t_orth (t_1, upperInit) | token t_1 begins with an initial uppercase letter |
| | t_morph_type(t_1, word) | token t_1 is has the morphological type *word* |
| **Syntactical features** | | |
| *POS and POS n-grams* | t_pos (t_1, nnp) | token t_1 is a singular proper noun |
| | t_gpos(t_1,nn) | token t_1 is a canonical noun (no plurals) |
| | t_bigPosBef (t_1, ....) | POS tag bigram before token t_1 |
| | t_bigPosAft (t_1, vbz-vbg) | POS tag bigram after token t_1 |
| | t_trigPosBef (t_1, ....) | POS tag trigram before token t_1 |
| | t_trigPosAft (t_1 vbz-vbg-dt ) | POS tag trigram after token t_1 |
| *Chunking analysis* | ck_hasHead(ck_1, t_1) | ck_1 has t_1 as its token head |
| | ck_hasType(ck_1, np) | ck_1 is a nominal chunk |
| | t_isHeadNP(t_1) | t_1 is the head token of a nominal chunk |
| | ck_dist_to_root(ck_n, near) | ck_n is near the main verb of the sentence |
| | t_ck_tag_type( t_1, np) | token t_1 has the chunking type *np* |
| **Semantic features** | t_ner(t_1, person) | t_1 was annotated by the NER as PERSON entity |
| Predefined corpus annotation types | t_type(t_1, person) | t_1 has the PERSON corpus type |
| | t_subtype(t_1, none) | t_1 has no subtype |
| | t_mtype(t_1, name) | t_1 is a named proper noun |
| **Structural features** | t_next (t_1, t_2) | token t_1 is followed by the token t_2 |
| | t_next_head (t_1, t_3) | head token t_1 is followed by head token t_3 |
| | ck_hasToken(ck_1, t_1) | t_1 is one the tokens in the chunk ck_1 |
| | ck_hasSucc(ck_1, ck_2) | ck_1 is followed by the chunk ck_2 |
| | t_hasDep (nn, t_2, t_1) | t_1 has a multi-word dependency with t_2 |
| | t_root (t_n) | t_n is the root (main verb) of the dependency tree |

**User-defined BK**. In OntoILPER, the user can specify any form of additional declarative knowledge to help the rule induction process. The predicates displayed in Fig. 6 were also integrated as BK into OntoILPER. These user-defined predicates consist in two intentional predicates that discretize numerical features, including *token length/2* and *chunk dist_to_root/2*: the first predicate categorizes the token length as *short, medium or long size*, while the second discretizes the distance (in number of tokens) between a chunk and the main verb (*root*) of the sentence. The numerical values in these predicates were adjusted manually. Such user-defined predicates intend to enable better rule generalizations.

```
% Token length type definition
length_type(short). length_type(medium). length_type(long).

tok_length(T, short) :- token(T), t_length(T, X), X =< 5.
tok_length(T, medium):- token(T), t_length(T, X), X > 5, X =< 15.
tok_length(T, long)  :- token(T), t_length(T, X), X > 15.

%  Chunking distance to the main verb
ck_dist_root(CK, near):-  ck_posRelPred(CK, X), X >= -3, X  =< 3.
ck_dist_root(CK, far) :-  ck_posRelPred(CK, X), ( ( X >= -8, X < -3) ;
                          (X > 3, X =< 8)).
ck_dist_root(CK, very_far):-  ck_posRelPred(CK, X),(( X < -8); (X > 8)).
```

**Figure. 6** Intentional predicates added to the original BK in OntoILPER.

### 3.4. ILP-based Rule Learning Component

The rule learning component in OntoILPER integrates ProGolem [Santos, 2010], [Muggleton et al., 2009], an efficient bottom-up ILP learner capable of learning complex non-determinate target predicates. ProGolem combines the most-specific clause construction of Progol [Muggleton, 1995] with the bottom-up control strategy of Golem [Muggleton and Feng, 1992]. It is based on the *predictive setting* which employs ILP for constructing classification models expressed as symbolic rules able to distinguish between positive and negative examples.

According to Santos (2010), ProGolem has an advantage with respect to top-down ILP systems, like Aleph, because ProGolem is able to learn long, non-determinate target concepts or predicates. In many real-world applications, such as the learning of chemical properties from atom and bond descriptions, the target predicate complexity is usually unknown a priori, i.e., it is problem dependent. Such complexity requires non-determinate BK as well. The basic covering set algorithm used in ProGolem (Santos, 2010) is given below:

```
ProGolem Covering Set Algorithm

Input:  Examples E, background knowledge B, mode declarations M
Output: Theory T, a set of definite clauses or rules
  1: T = {}
  2: E+ = all positive examples in E
  3: while E+ contains unseen positive examples do
  4:     e = first unseen positive example from E+
  5:     Mark e as seen
  6:     C  = best_armg(e, E, M)
  7:     Ce = negative_based_reduction(C, E)
  8:     if Ce has positive score then
  9:        T  := T ∪ Ce
 10:        E+c := all positive examples that clause Ce covers
 11:        E+ := E+ - E+c
 12:     end if
 13: end while
 14: return T
```

As shown above, the covering set algorithm is used by ProGolem in order to construct a theory consisting of possibly more than one clause. At each iteration of this algorithm, to select the highest-scoring armg of an initial seed example $e$ (line 4), ProGolem iteratively build new clauses calling the beam-search iterated ARMG (asymmetric relative minimal generalizations) algorithm (line 6) proposed in [Hitzler et al., 2009]. Then, the clauses found by the beam-search will be further generalized. For pruning the literals from the body of a given clause $C$, ProGolem employs a negative-based reduction strategy (line 7) that removes the non-essential literals, i.e., the ones that can be removed without changing the negative coverage of the clause $C$.

Finally, if the current clause $C_e$ achieves an expected accuracy score (line 8), it is added to the theory $T$ and all the examples covered by it are removed from the set of training examples. A detailed description of ARMG and negative-based reduction algorithms can be found in (Santos, 2010).

ProGolem imposes some restrictions over the induced extraction rules [Santos, 2010], as follows: (i) they have to reflect the BK in terms of both structural and property features defined by the relational model of sentence representation described in Section 3.2.; (ii) they must be well-formed with respect to the *linkedness* of the variables in the rules, i.e., there must exist a chain of literals connecting the input variables in the head of a rule to the variables in the body of the rules [Santos, 2010], (iii) they should define linguist patterns easily interpretable by the domain expert.

In what follows, the general aspects related to rule induction in OntoILPER are presented, followed by mode declarations and some examples of induced rules.

**Generating Extraction Rules.** During learning in OntoILPER, the search for rules in the hypothesis space that ProGolem has to perform is computationally expensive because it is necessary to test each candidate rules with respect to the positive and negative examples. Actually, this is the most expensive task in the entire ILP learning process. To speed up learning, ProGolem intelligently goes through the hypothesis space, taking advantage of its particular structure, only exploring the portions of the hypothesis space containing highly accuracy extraction rules. For that, the hypothesis space is structured by a quasi-order relation between two hypotheses which allows an efficient navigation among the candidate rules [Muggleton & Feng, 1992] [Santos, 2010]. In addition, ProGolem performs an efficient search for the most accurate extraction rules by delimiting and biasing the possibly huge hypothesis search space via mode declarations which are presented next.

**Mode Declarations.** Mode declaration [Muggleton & Feng, 1992] is one of the most known types of bias employed by ILP systems, including ProGolem, for defining syntactical constraints on the form of the valid rules. They also inform the type, and the input/output modes of the predicate arguments in a rule [Santos, 2010]. ProGolem provides two types of mode declarations: *head* and *body*. The former (*modeh*) denote the target predicate, in other words, the head portion of a valid rule to be induced, whereas the latter (*modeb*) constraints the literal or ground atoms that can appear in the body part of the rule. Mode declarations also impose restrictions on the types of the variables used as arguments of a predicate. Such types are simply declared by Prolog predicates of the form *type(value)*, e.g., *token(t_1)* and *chunck(ck_1)* which are used as identifiers of tokens and chunks, respectively.

The mode declarations corresponding to some of the features in Table 1 are listed below.

```
:- modeh(1, work_for(+token, +token)).      % Head or target predicate
:- modeb(*, t_hasDep(#dep, +token, -token)). % Structural
:- modeb(*, t_next(+token, -token)).
:- modeb(*, ck_has_tokens(-chunk, +token)).  % Chunking
:- modeb(*, ck_hasSucc(+chunk, -chunk)).
:- modeb(*, t_pos(+token,#postag)).          % Syntactic (POS)
:- modeb(*, t_trigPosBef(+token,#trigposbef)).
:- modeb(*, ck_hasType(+chunk, #ck_tag)).    % Chunking-related
:- modeb(*, ck_hasHead(+chunk, #token)).
:- modeb(*, t_ner(+token,#ner)).             % Semantic NER
```

At the beginning of a mode declaration definition the symbol "1" means that only one instance of the accompanying predicate can appear in the rule, while "*" means that any number of accompanying predicate can appear in the body part of the rule. For instance, the first mode declaration above denotes the head of the rule *work_for*, i.e., only one instance of the target predicate *work_for (token, token)* is allowed in the rule, denoting a binary relation between two tokens. The third mode declaration denotes the predicate *t_next(token,token)* that links a token to the next one in a sentence. Finally, the symbols "+" and "-" restrict the way a predicate (or literal) is followed by the previous one during rule learning. The interested reader is referred to [Santos, 2010] [Muggleton, 1995] for more information about mode declarations in ProGolem.

During the learning phase, mode declarations severely limit the number of potential solutions and ensure that only well-formed hypotheses are generated. In OntoILPER, a well-formed hypothesis is defined as a clause providing information about the entities, and the words appearing in their contexts.

**Induced Rules**. Fig. 7 illustrates two induced extraction rules for the *part_whole* relation. Such extractions rules are regarded as symbolic classifiers in OntoILPER and can be applied for classifying new unseen examples.

**Rule 1:**

  *part_w(A,B):- t_gpos(A,nn), t_next(A,B), t_subtype(B,state-or-province).*

**Rule 2:**

  *part_w(A,B):- t_next(A,B), t_pos(A,nnp), t_ne_type(B,gpl), t_subtype(A,pop-center).*

**Figure 7.** Two extraction rules for the *part_whole* relation induced from a news domain corpus.

*Rule 1* classifies an instance of the part-whole relation. Thus, sentences containing *two adjacent tokens (or phrases)* where the first one (*A*) is a *noun*, and the second one (*B*) is tagged with respect to the domain ontology as an instance of the "*State-or-Provence*" subtype class. This rule highlights that places (*A*) like cities are located, or are part of either a State or Provence.

*Rule 2* is very similar, as the entity instances indicated by the tokens variables *A* and *B* are also adjacent. In this rule, the token *A* is identified as a *proper noun* and is mapped to the *Population-Center* subclass in the domain ontology; whereas the token *B* is an instance of the named entity *Geographical Political Location* (*GPL*).

### 3.5. Rule Application Component

The goal of this component is to apply the induced rules in the knowledge base (Prolog factual base) generated from new documents similar to the ones used in the rule learning phase. As a result, new instances of relations are identified and extracted. The extracted instances are then added as new instances in the input domain ontology. This task is also known as Ontology Population. In fact, OntoILPER can be regarded as an ontology population system. The OWL API[7] was used for implementing the ontology population service.

Due to its importance as a core component in OntoILPER, the roles played by the domain ontology are detailed next.

### 3.6. The Role of the Domain Ontology

In OntoILPER, the domain ontology plays two important roles in (i) *training mode* and (ii) *application mode*.

**Training Mode**. The domain ontology describes both the domain and the BK exploited by OntoILPER. This ontology guides the BK generation process by defining the level of abstraction (classes and superclasses) of the BK predicates from which the rules will be induced. Therefore, TBox elements of the domain ontology (class and property labels, data/object properties, taxonomical relationships, and domain/range of non-taxonomical relations) are taken into account during the BK Generation step. In other words, the domain ontology integration into the OntoILPER IE process conforms to the first three levels of ontological knowledge used by some of the state-of-the-art OBIE systems, as discussed in [Karkeletis, 2011]: At the first level, the ontological resources explored by OntoILPER encompass the domain entities (e.g., person, location) and their synonyms. These resources are mainly used in OntoILPER for entity classification. At the second level, the domain entities are organized in conceptual hierarchies, which can be exploited by OntoILPER IE process for generalizing/specializing extraction rules. At the third level, OntoILPER exploits both the properties of the concepts and the relations between concepts of the ontology. Moreover, extraction rules are acquired from corpora that have been previously annotated according to the domain ontology. Finally, the domain ontology can be viewed as a structured extraction template for the IE process.

**Application Mode**. OntoILPER overall RE process aims at mapping pieces of textual information to the domain ontology. OntoILPER selects and interprets relevant pieces of the input text in terms of their corresponding classes in the domain ontology. Fig. 8 illustrates this semantic mapping process.

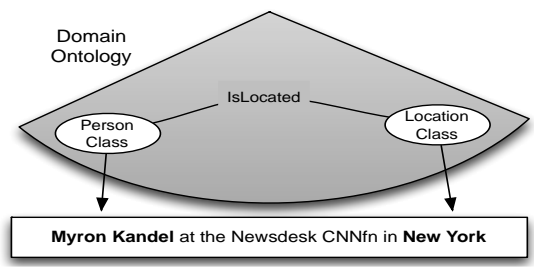

**Figure 8**. An example of semantic mapping.

## 4. Experimental Evaluation

This section reports and discusses the results of comparative experiments among state-of-the-art supervised RE systems and OntoILPER, conducted on three datasets from the biomedical domain concerning protein-protein interactions. In what follows, both the evaluation datasets and measures used in the experiments are first described. Then, we present the evaluation methods, cross-validation and cross-corpus, as well as our experimental setup aiming at comparing OntoILPER performance on RE with related work.

### 4.1 Datasets and Evaluation Measures

---

[7] The OWL API. http://owlapi.sourceforge.net

**Datasets.** Three publicly available datasets from the biomedical domain containing PPI [Airola et al., 2008], i.e., binary relations between two proteins, were selected:

- *Learning Language in Logic* (LLL) [Nedellec, 2005]. This dataset introduces the gene interaction task from a set of sentences concerning *Bacillus subtilis* transcription.
- *HPRD50* [Fundel et al., 2007]. It consists of a randomly selected subset of 50 abstracts referenced by the Human Protein Reference Database (HPRD).
- *Interaction Extraction Performance Assessment* (IEPA) [Ding et al., 2002]. This dataset is composed of 303 abstracts taken from the PubMed repository, each one containing a specific pair of co-occurring chemicals.

We cast the PPI extraction task as a binary classification problem, in which interacting protein pairs are considered positive examples and the other co-occurring pairs in the same sentence are negative ones. It is worth noticing that the interactions are marked at the level of proteins (entity) pairs, which allows for the annotation of multiple entity pairs per sentence.

Although all of the datasets provide annotations about named biomedical entities and protein-protein interactions, there are many aspects in which these datasets differ significantly. For example, they differ in the scope of the annotated entities varying from proteins to genes, RNA and chemicals [Tikk, 2012]. In addition, the coverage of the annotated entities is not complete, and some datasets specify the direction of interactions and others do not [Psysalo et al., 2008]. Tab. 3 summarizes the basic statistics of the PPI datasets concerning the number of sentences (#Sentences), and the number of positive (#E$^+$) and negative (#E$^-$) examples.

**Table 3.** Basic statistics of the three PPI datasets.

| Corpus | #Sentences | #E+ | #E- |
|---|---|---|---|
| LLL | 77 | 164 | 166 |
| HPRD50 | 145 | 163 | 270 |
| IEPA | 486 | 335 | 482 |

**Evaluation Measures.** We use the classical IR measures of Precision *P*, Recall *R*, and F1-measure [Baeza-Yates and Ribeiro-Neto, 1999] for assessing the effectiveness of the evaluated systems in this section.

F1-measure has been criticized as inadequate for PPI extraction due to its sensitivity to the unbalanced number of positive and negative examples in training set [Pyysalo et al., 2008]. For that reason, we also report on the results in terms of the Area under Curve (AUC) measure, as suggested in [Airola et al, 2008]. Furthermore, in order to have a fair comparison with previously published results on the same PPI datasets and shared tasks, we have adopted the same performance criteria and experimental setup proposed in [Tikk et al., 2010] and [Airola et al, 2008] and also adopted by the other RE systems cited in this section, namely: protein-protein interactions are considered as both untyped and symmetric pairs of mentions; self-interactions are removed from the corpora before evaluation; and gold standard protein (entities) annotations are used as they are provided by the corpora.

### 4.2 Optimal ProGolem Parameters

For generating models with high classification accuracy, it is first necessary to find optimal learning parameters. To obtain such parameters for ProGolem, we followed the method proposed in [Kohavi and John, 1995] which recommends the following three steps: (i) separating the most relevant parameters; (ii) obtaining unbiased estimates of the classification accuracy of the models built after a systematic variation across some small number of values for the parameters chosen in the previous step; (iii) taking the values that yielded the best average predictive accuracy across all target predicates.

Accordingly, several preliminary experiments were performed for achieving both high accuracy results and preventing model overfitting. As a result, the following optimal learning parameters of the ProGolem algorithm were stabilised: {*theory_construction = global, evalfn = coverage, depth_i = 3, minprec = 0.5, minpos = 3, noise = 0.3*}. The last three parameters avoid overfitting by imposing that an extraction rule must cover a minimum number of positive examples (minpos); and achieve a minimum precision (minprec). Finally, the noise parameter allows for selecting rules covering only few negative examples.

### 4.3 Experimental Setup

*4.3.1. Evaluation Settings: Cross-Validation and Cross-Corpus*

Our experimental setup is composed of two evaluation settings: *Cross-Validation* (CV) and *Cross-Corpus* (CC) [Airola et al., 2008, Tikk et al., 2010]. In CV, we train and test OntoILPER and the selected RE kernels on the same corpus using 10-fold cross-validation because it allows both fairly maximal use of the available data and direct performance comparison of the results. Although CV is still the current de facto standard in PPI extraction, it can also provide somewhat biased performance estimations due to the fact that both training and test examples usually share very similar characteristics [Airola et al., 2008].

In order to both minimize such biased results, we also conduct Cross-Corpus (CC) evaluation in which the training and test datasets are drawn from different distributions, i.e., we train the RE systems on one corpus and test them on other corpora.

We are particularly interested in CC evaluation in OntoILPER for two reasons: it may confirm the existence of overlapping extraction rules or patterns among the corpora, and it enables the assessment of the generalization level of the final rules generated by OntoILPER. Furthermore, as already mentioned by Airola et al. (2008), CC evaluation concerns a research question of paramount importance when training OntoILPER: *can the induced extraction rules generalize beyond the specific characteristics of the data used for training them?*

*4.3.1 Selected RE Systems for Comparison*

We selected six RE systems to compare with OntoILPER on same biomedical datasets described in Section 4.1. These systems were selected because they achieved the best performance on PPI extraction as reported in [Tikk et al., 2010]. All of the kernel-based RE systems used the SVM algorithm and were evaluated using both CV and CC methods and employed the same experimental settings as ours. We rather selected the kernel-based RE systems instead of the ILP-based RE systems (cf. Section 2.2) more closely related to OntoILPER. The reason is that, on the one hand, the kernel-based RE systems employed the same experimental setting and the same publicly available PPI extraction datasets for RE, which allows for a direct and fair comparison with OntoILPER. On the other hand, almost all of the ILP-based RE systems were evaluated using tailored or non-publicly available corpora, therefore a direct comparison among them and OntoILPER is not possible. The selected RE systems are shortly described next.

- **SL [Giuliano et al., 2006]**. The Shallow Linguistic kernel (SL) is exclusively based on two types of contextual features: local and global. The former is based on surface features including punctuation, capitalization; the latter is based on shallow linguistic features such as lemma, and POS tagging in the w size window of the entities.

- **APG [Airola et al., 2008].** The All-Path Graph kernel (APG) counts weighted shared paths of all possible lengths in the parsing tree. Path weights are determined by the distances of the dependencies as the shortest path between the candidate entities in a relation. It uses as features the dependency graph with POS tags, word sequence, and weighted edges along the shortest path in the dependency graph.

- **kBSPS [Tikk et al., 2010].** The k-band Shortest Path Spectrum kernel (kBPSS) combines three others base kernels: the syntax-tree which is adapted to dependency trees; the vertex-walk based on the sequence of edge-connected tree nodes; and a variant of the APG that also take into account neighboring nodes.

- **Composite [Miwa et al. 2009, 2010].** The AkaneRE RE system [Miwa et al. (2009)] is based on a composite kernel combining three others: a bag-of-words kernel that calculates the similarity between two unordered sets of words; a subset tree kernel that calculates the similarity between two input trees by counting the number of common subtrees; and a graph kernel that calculates the similarity between two input graphs by comparing the relations between common nodes. An improved version of AkaneRE [Miwa et al., 2010] applies a set of simplification rules over the output of its parser for removing unnecessary information from the parse trees.

- **SDP-CPT [Quian & Zhou, 2012].** The Shortest Dependency Path-directed Constituent Parse Tree (SDP-CPT) kernel uses the shortest dependency path between two proteins in the dependency graph structure of a sentence for reshaping the constituent parse tree.

- **EOEP-CPT [Ma et al., 2015].** The Effective Optimization and Expanding Path-directed Constituent Parse Tree (EOEP-CPT) [Ma et al., 2015] kernel is an improvement over the SDP-CPT kernel. This kernel corrects parsing errors by applying processing rules that can optimize and expand the shortest dependency path between two proteins in a given sentence.

*4.3.2 Comparative Cross-Validation Evaluation*

Tab. 4 summarizes the results where the highest scores for F1-measure are in **bold**, and the best AUC scores in *italics*.

**Table 4.** Cross-validation results of the RE systems on the PPI corpora

| Corpus | OntoILPER | | APG [Airola et al., 2008] | | kBSPS [Tikk et al., 2010] | | Composite [Miwa et al., 2010] | | SDP-CPT [Quian & Zhou, 2012] | | EOEP-CPT [Ma et al., 2015] | |
|---|---|---|---|---|---|---|---|---|---|---|---|---|
| | **F1** | **AUC** | **F1** | **AUC** | **F1** | **AUC** | **F1** | **AUC** | **F1** | **AUC** | **F1** | **AUC** |
| LLL | 79.9 | 85.2 | 76.8 | 83.4 | 78.1 | 84.3 | 82.9 | *90.5* | **84.6** | 89.9 | 82.3 | 87.2 |
| HPRD50 | **75.3** | *87.6* | 63.4 | 79.7 | 71.0 | 79.3 | 75.0 | 86.6 | 68.8 | 83.7 | 65.1 | 81.9 |
| IEPA | 76.1 | 87.2 | 75.1 | 85.1 | 70.5 | 83.2 | **77.8** | *88.7* | 69.8 | 82.8 | 68.7 | 81.6 |
| *average* | 77.10 | 86.67 | 71.77 | 82.73 | 73.20 | 82.27 | **78.57** | *88.60* | 74.40 | 85.47 | 72.03 | 83.57 |

The overall results in Tab. 4 show that SDP-CPT had the best performance on the LLL dataset in terms of F1-measure. For this same dataset, the highest AUC was obtained by the Composite kernel method. OntoILPER outperformed, in terms of both F1-measure and AUC scores, all other systems on the HPRD50 dataset. For the IEPA dataset, the Composite kernel obtained the highest F1 and AUC scores, with OntoILPER coming just behind it.

The SDP-CPT kernel results in Tab. 4 were taken from Quian and Zhou (2012). However, in [Ma et al., 2015], the authors reported an improvement of 1,2% percentage points in F1 over the SDP-CPT kernel results with their enhanced EOEP-CPT kernel. The main reason for such an improvement, according to the authors, is due to the fact that EOEP-CPT algorithm is more precise and concise than the SDP-CPT when both kernels uses the same version of the Stanford parser (v2.0.4). Furthermore, Ma and colleagues claim that the EOEP-CPT algorithm is effective in removing noise interference caused by appositive dependency relation, while retaining critical information.

Among all the compared systems in Tab. 4, the Composite kernel is the only one that performs a simplification step similar to OntoILPER. By the way, the Composite kernel version equipped with simplification rules improved F1-measure in almost 4 points compared to its previous version reported in [Miwa et al., 2009]. Moreover, although the Composite kernel had obtained the overall highest scores on the IEPA corpus, this kernel has high computation complexity and difficulty in implementation [Ma et al., 2015].

As already mentioned in Section 3.2, OntoILPER graph-based transformation method also performs the simplification of the graphs representing sentences. This simplification approach contributes to the overall boost in performance (mainly in terms of recall) of the final induced extraction rules, as reported in a previous work [Lima et al., 2014a]. This is also evidenced in this work in which OntoILPER achieved the second best overall averaged performance on all of the PPI datasets.

OntoILPER simplification method differs from the one proposed in [Miwa et al., 2010] mainly due to the nature of the simplification rules, and the target syntactical constructions of the sentences. In fact, OntoILPER relies on typed dependencies originated by a dependency parser, whereas Miwa and colleagues' system uses a constituent parser. As a result, OntoILPER simplification rules tend to be simpler and more flexible with respect to the position order of both the target entity and the main clause in the sentence. On the other hand, contrarily to the extraction rules based on the output of a dependency parser, extraction rules based on constituent parsers have to consider the exact position of the target elements in a sentence [Buyko et al., 2011].

### 4.3.3 Comparative Cross-Corpus Evaluation

Tab. 5 summarizes OntoILPER CC evaluation results with those reported in [Tikk et al., 2010] and [Ma et al., 2015] for the selected PPI extraction kernel systems. It was adopted the same experimental setup used by the other systems in order to have a fair comparison. The main goal of this experiment is to answer the question about the capability of OntoILPER to learn rules that generalize beyond the specific characteristics of the data used for training it.

It is worth mentioning that the kernel-based solutions shown in Tab. 5 are mostly convolution kernels, i.e., they exploit the structure of the examples, i.e., syntax trees or dependency graphs of sentences. Their main idea is to calculate the overall similarity of two given examples through calculating the similarities of their substructures.

**Table 5.** Cross-corpus results (F1-score). Rows correspond to training corpora and columns to test corpora. Bold values denote the highest scores in CC among the methods[8]. Cross-validation results are indicated in italics.

|   | OntoILPER | | | SL [Giuliano et al., 2006] | | | APG [Airola et al., 2008] | | | kBSPS [Tikk et al., 2010] | | | EOEP-CPT [Ma et al., 2015] | | |
|---|---|---|---|---|---|---|---|---|---|---|---|---|---|---|---|
|   | L | H | I | L | H | I | L | H | I | L | H | I | L | H | I |
| L | *79.9* | 61.8 | 68.0 | *74.5* | 59.0 | 62.6 | 76.8 | 63.0 | 63.0 | 78.1 | 62.6 | 66.6 | 82.3 | **64.2** | **68.1** |
| H | 55.9 | *75.3* | 57.4 | 61.4 | *64.2* | 53.3 | **62.1** | 63.4 | **60.9** | 62.0 | *71.0* | 55.7 | 54.2 | *65.1* | 54.2 |
| I | 59.5 | 54.1 | *76.1* | 66.4 | 58.2 | *69.3* | 61.8 | 63.9 | *75.1* | **75.9** | 64.3 | *70.5* | 67.8 | **66.3** | *68.7* |

Datasets: L = LLL, H = HPRD50, I = IEPA

The CC results summarized in Tab. 5 show some performance gaps for most of the kernel-based systems and OntoILPER depending on the training corpus used. On average, the systems trained on the LLL dataset performed better.

On the one hand, OntoILPER models trained on the LLL dataset performed better than the majority of the other RE systems. On the other hand, APG achieved the highest F1-scores in two pairs of training/testing datasets (HPPDR/LLL and HPPDR/LLL). This result can be explained by the fact that the APG kernel was specifically designed by taking into account the particular characteristics of the HPPDR dataset. The kBSPS kernel was significantly superior than the others systems in only one pair of the PPI datasets (IEPA/LLL), while the EOEP-CPT kernel showed superior performance over the others for only two pairs of training/testing datasets, as the difference between EOEP-CPT and OntoILPER for the LLL/IEPA (training/test) dataset pair was minimal.

Overall, such results suggest that there is no a clear winner among the compared systems in the CC evaluation. Indeed, several Wilcoxon signed-rank statistical tests were performed on the results of Table 7 to check statistical significance in the difference in performance between each pair of the compared systems. The test results showed that there is no significant difference in performance among all of them at 95% confidence level ($\alpha = 0.05$).

**Discussion**. The performance of the systems for PPI extraction shown on Tab. 5 clearly depends on the differences between the training data and the test data. Such differences consist in distinct writing styles, the level of technical detail or the frequency of certain linguistic grammatical structures, among others.

The conclusion is that, due to the different characteristics present in the training and the test datasets, one can generally expect lower performance in CC than in CV evaluation. Moreover, these results also indicate that none dataset is a bad choice for training. Indeed, further investigation is still needed to better understand the impact of the singular characteristics of the datasets under analysis concerning RE systems in general.

The overall results shown in Tab. 5 suggests that OntoILPER approach is comparable with kernel-based RE systems in the sense that the underlying structures of the examples, given by the relational model of sentences, are also able to find the most important features of the examples as the convolution kernels can do. In other words, the relational model of examples which takes into account both the lexical and syntactical features, represented as directed connected graphs, is able to capture the most important structural information from the examples without leaving behind major relational features.

We also highlight that most of the compared RE kernels discussed this section have been specifically designed to the PPI extraction task, i.e., they take into account the particularities of the sentence structures or the style preferences found in the biomedical corpora. This contrasts with OntoILPER which is domain independent. In fact, OntoILPER has achieved state-of-the-art performance on the News Broadcast domain as well, as reported in [Lima et al., 2014b]. Therefore, a direct way of trying to improve OntoILPER performance on PPI extraction is to enable it with a specific named entity recognizer for biomedical entities as well as using parsers trained on biomedical domain datasets. Finally, the encouraging results of OntoILPER put it as a valuable alternative RE method which presents several advantages over the RE methods based on kernels.

## 4.4 OntoILPER Limitations

---

[8] CC evaluation results on the three datasets in Table 7 were not provided for the Composite and SDP-CPT kernels because they were not published in their respective papers.

In the following, we discuss some OntoILPER drawbacks along with some strategies to overcome or alleviate them.

**Parsing Errors in Text Preprocessing.** OntoILPER adopts the pipelined architecture commonly used by NLP-based applications. This architecture is well known to be prone to parsing errors which might hamper extraction performance due to the noisy patterns introduced in the relational representation of the sentences. Actually, for some domains, especially for the biomedical one, the complexity of sentences poses a challenge to natural language parsers, which are commonly trained on large-scale corpora composed of non-technical text (news broadcast texts).

In order to alleviate parsing errors, as reported in [Pyysalo et al. 2007] and [Jonnalagadda et al., 2009], sentence simplification techniques can substantially improve the average accuracy of natural language parsers in more than 4% in many datasets from different domains. Such finding has inspired our investigation to apply a similar method for sentence simplification in OntoILPER (cf. Section 3.2.2) which obtained very similar results discussed in [Lima et al., 2014a].

**High Computation Cost of Learning.** The searching process for high accurate extraction rules in OntoILPER is time-consuming, mainly due to the covering algorithm employed by its ILP system that generates and tests hundreds or even thousands of candidate extraction rules during learning. To alleviate that, undersampling techniques [Chawla, 2005] [Liu et al., 2009] which consists in reducing the number of examples (mainly the negative ones) should be explored. Undersampling techniques select the training examples, reducing the hypothesis space to a more manageable size. Combined with the above graph-based reduction of sentence representation techniques, sample size selection methods [Byrd et al., 2012] will also be investigated.

**Overfitting of the Extraction Rules.** Machine learning algorithms often tend to overfit (memorize) their training data. The ILP approach adopted in OntoILPER is biased toward producing highly precise, but low recall extraction rules. In fact, by inspecting the extraction rules of the experiments reported in Section 4, it was found that 10-20% of the extraction rules were quite specialized, i.e., covering at most two distinct positive examples. For tackling this problem, ensemble machine learning methods [Goadrich et al., 2006][Dietterich, 2000] could be of help.

## 5. Conclusion and Future Work

This work presented OntoILPER, an ILP-based method for extracting instances of binary relations from texts that exploits a domain ontology in its extraction process. OntoILPER is composed of two main components: an effective graph-based model for sentence representation and an ILP-based learning component. The former defines a hypothesis space that enables the searching for candidate hypothesis space, whereas the latter relies on ILP to induce Horn-like extraction rules from annotated examples.

The study of related work has shown that that most of the existing state-of-the-art approaches to RE are kernel-based. Contrary to this trend, this paper discussed the main advantages of an ILP-based learning component that allows for integrating additional BK in many ways, notably by exploiting a domain ontology given as input.

OntoILPER was evaluated using three golden standard datasets on the biomedical domain. Experiments conducted on these datasets allowed a comparative assessment of OntoILPER with other state-of-the-art RE systems. OntoILPER achieved encouraging results that demonstrated its effectiveness; however, there is still room for improvement.

OntoILPER is currently based on the shallow syntactic parsing of sentences, which does not take into account semantic aspects relating entities to verbs. Accordingly, we intend to integrate further BK (semantic resources) into OntoILPER preprocessing stage, such as synonyms, hypernyms/hyponyms, SRL [Christensen et al., 2010], and word sense disambiguation [Ciaramita and Altun, 2006], since these semantic resources have recently been proven to improve performance in many IE applications [Dou et al., 2015]. In this case, the distinctive feature of ILP would allow incremental BK to be put to test. Our long-term goal is to fully exploit different types of BK aiming at investigating which kind of BK is more useful to specific domains.

We will also investigate ILP-based rule induction from larger datasets aiming to promote OntoILPER scalability. Such improvement is possible, for example, by sampling techniques, for selecting the most informative examples and removing the redundant ones [Byrd et al., 2012]; and parallel ILP learning processing [Camacho et al., 2014, 2016] [Srinivasan et al., 2012] that can decompose the learning problem into smaller more manageable parts. Another future line of study concerns adapting OntoILPER for performing Event Extraction that aims at identifying *n*-ary relations in the biomedical domain [Björne and Salakoski, 2015].

## Acknowledgment

The authors would like to thank Hilario Oliveira for helping us in the implementation of the Text Preprocessing and BK Generation modules, and the National Council for Scientific and Technological Development (CNPq/Brazil) for financial support (Grant No. 140791/2010-8).